%% file: lrgwn.tex
\documentclass{article}

    \PassOptionsToPackage{numbers, compress}{natbib}

\input{sections/header/style_format}
\input{math_commands}

\usepackage[utf8]{inputenc}         
\usepackage[T1]{fontenc}            
\usepackage{bookmark}
\usepackage{xkcdcolors}
\usepackage{booktabs}               
\usepackage{nicefrac}               
\usepackage{microtype}              
\usepackage{xcolor}                 

\usepackage{amsthm}
\usepackage{amssymb}
\usepackage{wrapfig}
\usepackage{graphicx}
\usepackage{multirow}
\usepackage{cuted}
\usepackage[font=small]{caption}
\usepackage{subcaption}
\usepackage[capitalize]{cleveref}
\usepackage[many]{tcolorbox}
\usepackage{pgfplots}

\pgfplotsset{compat=1.18}

\newcommand{\modelname}{LR-GWN}
\newtheorem{theorem}{Theorem}
\newtheorem{lemma}{Lemma}
\newtheorem{proposition}{Proposition}
\newtheorem{definition}{Definition}

\PassOptionsToPackage{pdftex,pdfusetitle}{hyperref}
\hypersetup{
    pdftitle={Long-Range Graph Wavelet Networks},
    colorlinks=true,
    linkcolor=xkcdCrimson,
    urlcolor=xkcdBluish,
    citecolor=xkcdLightNavy,
}

\tcbset{
  insightbox/.style={
    colback=gray!5,          
    colframe=black!30,       
    boxrule=0.4pt,           
    left=4pt,
    right=4pt,
    top=4pt,
    bottom=4pt,
    before skip=5pt,
    after skip=5pt,
    enhanced,                
  }
}

\input{sections/header/authors}

\begin{document}

\maketitle
\input{sections/0-abstract}
\input{sections/1-introduction}
\input{sections/2-background}
\input{sections/3-polynomial-filters}
\input{sections/4-method}
\input{sections/5-results}
\input{sections/6-discussion}
\input{sections/99-acknowledgments}



\bibliographystyle{iclr_2026}
\bibliography{references}

\clearpage


\appendix
\bookmarksetupnext{level=part}
\pdfbookmark{Appendix}{appendix}
\input{sections/appendix/extended_background}
\input{sections/appendix/polynomial_filter_limitations}
\input{sections/appendix/related_work}

\input{sections/appendix/implementation}

\input{sections/appendix/additional_results}
\input{sections/appendix/proofs}


\end{document}

%% file: sections/header/style_format.tex
 \usepackage[dblblindworkshop, final]{neurips_2025}
\workshoptitle{New Perspectives in Advancing Graph Machine Learning}

\title{Long-Range Graph Wavelet Networks}

%% file: math_commands.tex
\usepackage{amsmath,amsfonts,bm}

\def\eqref#1{equation~\ref{#1}}

\def\1{\bm{1}}

\def\vx{{\bm{x}}}

\def\mA{{\bm{A}}}

\def\mD{{\bm{D}}}

\def\mH{{\bm{H}}}
\def\mI{{\bm{I}}}

\def\mL{{\bm{L}}}

\def\mU{{\bm{U}}}

\def\mW{{\bm{W}}}
\def\mX{{\bm{X}}}

\def\mLambda{{\bm{\Lambda}}}

\DeclareMathAlphabet{\mathsfit}{\encodingdefault}{\sfdefault}{m}{sl}
\SetMathAlphabet{\mathsfit}{bold}{\encodingdefault}{\sfdefault}{bx}{n}

\def\gG{{\mathcal{G}}}

\def\gL{{\mathcal{L}}}

\def\gO{{\mathcal{O}}}

\newcommand{\R}{\mathbb{R}}



%% file: sections/header/authors.tex
\author{%
  Filippo Guerranti\\
  \And
  Fabrizio Forte\\
  \And
  Simon Geisler\thanks{Now at Google Research.}\\
  \And
  Stephan G{\"u}nnemann\\
  \AND
  \vspace{-2em}\\
  {\small \texttt{\{f.guerranti,f.forte,s.geisler,s.guennemann\}@tum.de}} \\
  {\small School of Computation, Information and Technology \& Munich Data Science Institute}\\
  {\small Technical University of Munich, Germany} \\
}

%% file: sections/0-abstract.tex
\begin{abstract}

Modeling long-range interactions, the propagation of information across distant parts of a graph, is a central challenge in graph machine learning. Graph wavelets, inspired by multi-resolution signal processing, provide a principled way to capture both local and global structures. However, existing wavelet-based graph neural networks rely on finite-order polynomial approximations, which limit their receptive fields and hinder long-range propagation. We propose Long-Range Graph Wavelet Networks (\modelname{}), which decompose wavelet filters into complementary local and global components. Local aggregation is handled with efficient low-order polynomials, while long-range interactions are captured through a flexible spectral-domain parameterization. This hybrid design unifies short- and long-distance information flow within a principled wavelet framework. Experiments show that \modelname{} achieves state-of-the-art performance among wavelet-based methods on long-range benchmarks, while remaining competitive on short-range datasets.

\end{abstract}

%% file: sections/1-introduction.tex
\section{Introduction}
\label{sec:intro}

Long-range interactions are central to complex systems, from electron correlations in quantum chemistry \citep{ambrosetti_long-range_2014, knorzer_long-range_2022} to allosteric effects in biology \citep{dokholyan_controlling_2016, zhu_neural_2022}. In graph neural networks (GNNs) \citep{gori_new_2005, scarselli_graph_2009, gilmer_neural_2017, bronstein_geometric_2017}, capturing these interactions requires models that can propagate information beyond local neighborhoods without the computational overhead of dense global interactions \citep{alon_bottleneck_2021, dwivedi_long_2022}.

Wavelet-based graph neural networks (WGNNs) \citep{hammond_wavelets_2011}, inspired by wavelet theory \citep{mallat_multiresolution_1989}, provide a principled framework for this challenge, defining spectral filters that can in principle capture different scales of information propagation through their filtering characteristics. However, designing exact wavelet filters face a fundamental computational bottleneck, as it would require computing the full eigenvalue decomposition of the graph Laplacian, which is prohibitive for large graphs.

To circumvent this cost, current WGNNs \citep{hammond_wavelets_2011, xu_graph_2019, liu_advancing_2024} approximate wavelet filters using low-order polynomials. While computationally efficient, this approach suffers from two critical limitations. First, polynomial filters aggregate information only within a finite number of hops \citep{balcilar_analyzing_2020}, inherently restricting their spatial reach. More fundamentally, polynomial approximations face an inherent trade-off between computational efficiency and functional expressiveness: while high-degree polynomials can theoretically approximate any function defined on the real interval (Weierstrass theorem), practically feasible orders cannot accurately represent the discontinuous or steep characteristics needed for selective frequency filtering, such as wavelets \citep{geisler2024spatiospectral}.

\input{figures/figure1}

Achieving both computational efficiency and sufficient functional expressiveness requires rethinking wavelet filter parametrization. Current approaches fail to resolve this trade-off: low-order polynomials confine propagation to local neighborhoods (\cref{fig:figure1}a), while higher orders extend the radius at additional computational cost, but still fail to achieve global propagation (\cref{fig:figure1}b). In contrast, the exact solution via full EVD enables global information flow but at prohibitive cost (\cref{fig:figure1}c).

We propose Long-Range Graph Wavelet Networks (\modelname{}), which overcome the limitations of polynomial-based WGNNs through a hybrid filter design. Our method decomposes each wavelet filter into two components: a low-order polynomial for efficient local aggregation and a spectral filter operating on the low-frequency eigenspace to enable long-range propagation (\cref{fig:figure1}d). This approach requires only a partial eigendecomposition at preprocessing time, making it computationally practical while achieving the global reach that purely polynomial methods cannot provide. 

\modelname{} can operate under strict wavelet theory or with relaxed constraints for enhanced performance, providing a principled yet flexible framework for wavelet-based long-range graph learning.

\textbf{Our contributions} are: 
(i) a \textbf{hybrid parametrization} combining polynomial spatial filters with learnable spectral filters operating on truncated eigenspaces;
(ii) a \textbf{principled wavelet framework} that maintains theoretical admissibility constraints while optionally allowing relaxation when needed; 
(iii) an \textbf{efficient implementation} requiring only partial EVD, adding minimal preprocessing overhead; and 
(iv) \textbf{state-of-the-art performance} among wavelet-based GNNs on long-range benchmarks.

%% file: figures/figure1.tex
\begin{figure*}[!t]
    \centering
    \includegraphics[width=\textwidth]{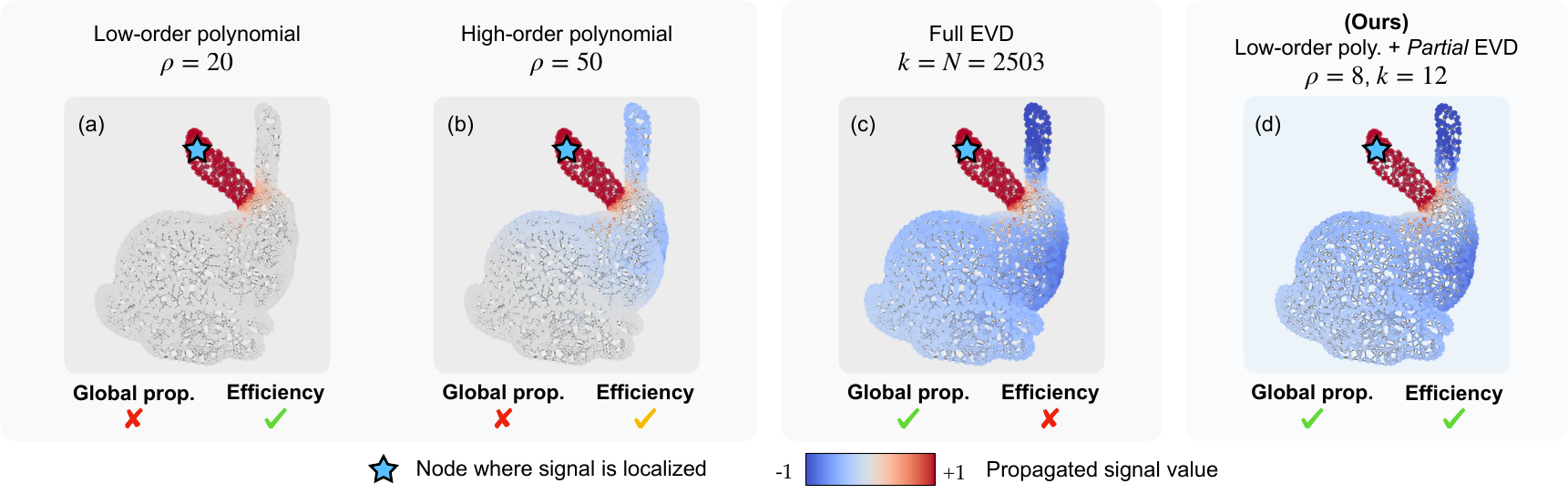}
    \caption{
       \textit{Signal propagation under different Mexican hat wavelet filter approximations.}  
       (a) Low-order polynomial ($\rho = 20$) restricts propagation to local neighborhoods.  
       (b) Higher-order polynomial ($\rho = 50$) extends reach but remains spatially bounded.  
       (c) Full EVD ($k = N = 2503$) enables global propagation at prohibitive cost.  
       (d) \modelname{} ($\rho = 8, k = 12$) achieves global propagation with minimal computational overhead.  
    }
    \label{fig:figure1}
\end{figure*}

%% file: sections/2-background.tex
\section{Background}
\label{sec:bg}

We recall the main tools from spectral graph theory and wavelet analysis; we refer the reader to \cref{app:extended_bg} for a detailed exposition on these topics.

\textbf{Graphs.}
We consider undirected graphs \(\smash{\gG=(\mA,\mX)}\) with adjacency matrix \(\smash{\mA\in\{0,1\}^{n\times n}}\) and node features \(\smash{\mX\in\R^{n\times d}}\). Throughout we use the symmetrically normalized Laplacian \(\smash{\gL=\mI-\mD^{-\frac12}\mA\mD^{-\frac12}}\), with \(\smash{\mD=\mathrm{diag}(\mA\bm{1})}\), which admits eigendecomposition \(\smash{\gL=\mU\mLambda\mU^\top}\) with orthogonal eigenvectors \(\mU \in \R^{n \times n}\) and eigenvalues \(\smash{\mLambda = \mathrm{diag}(\lambda_1,\dots,\lambda_n)} \in \R^{n \times n}\), \(\smash{0=\lambda_1\le\dots\le\lambda_n\le2}\) \citep{chung_spectral_1997}.
The eigenvectors of \(\smash{\gL}\) form an orthogonal basis, often called the graph Fourier basis. Any graph signal \(\smash{\vx\in\R^n}\) can be decomposed in this basis via the graph Fourier transform (GFT) as \(\smash{\widehat{\vx}=\mU^\top\vx}\), with inverse GFT \(\smash{\vx=\mU\widehat{\vx}}\). Spectral filtering applies a kernel \(\smash{g:\R\to\R}\) by modulating its Fourier coefficients: \(\smash{g\ast_{\gG}\vx=\mU\,\widehat g(\mLambda)\,\mU^\top\vx}\), where \(\smash{\widehat g(\mLambda)=\mathrm{diag}(\widehat g(\lambda_1),\dots,\widehat g(\lambda_n))}\). Small eigenvalues correspond to smooth/global variations, large ones to oscillatory/local patterns \citep[see Fig. 3 for an intuitive visualization]{kreuzer_rethinking_2021}.

\textbf{Wavelets.}  
Wavelet transforms extend classic Fourier analysis by providing joint localization in both frequency and space \citep{mallat_multiresolution_1989, mallat_wavelet_2009}. A mother wavelet \(\psi\) generates dilated and translated functions \(\smash{\psi_{s,l}(t)=s^{-1/2}\psi\!\left(\tfrac{t-l}{s}\right)}\), whose coefficients \(\smash{W_f(s,l)=\int_{\R} f(t)\psi^*_{s,l}(t)dt}\) decompose a signal \(f\) across scales \(s\) and locations \(l\). Reconstruction requires the admissibility condition, \(\smash{\widehat\psi(0)=0}\), ensuring \(\smash{\widehat\psi}\) acts as a band-pass filter. A complementary scaling function \(\phi\) with \(\smash{\widehat\phi(0)>0}\) provides low-pass coverage, yielding a multiresolution analysis where wavelets capture localized variations and the scaling function captures global trends.

\textbf{Wavelets on graphs.}  
The spectral graph wavelet transform (SGWT) \citep{hammond_wavelets_2011} adapts wavelet analysis to graphs using the spectrum of \(\smash{\gL}\). Since graphs lack translation and scaling, wavelets are defined spectrally: a band-pass kernel \(\smash{\widehat\psi}\) yields the wavelet operator \(\smash{\Psi=\mU\,\widehat\psi(\mLambda)\mU^\top}\), and dilation to scale \(s\) is obtained by \(\smash{\widehat\psi(s\mLambda)}\). A complementary scaling function defines \(\smash{\Phi=\mU\,\widehat\phi(\mLambda)\mU^\top}\). For a signal \(\smash{\vx}\), the wavelet coefficients at scale \(s\) are \(\smash{W_\vx(s)=\mU\,\widehat\psi(s\mLambda)\mU^\top\vx}\), while scaling coefficients are \(\smash{H_\vx=\mU\,\widehat\phi(\mLambda)\mU^\top\vx}\). Together, \(\smash{W_\vx(s)}\) and \(H_\vx\) provide a multiscale representation, where \(\smash{\widehat\psi}\) captures localized variations (band-pass filter) and \(\smash{\widehat\phi}\) global structure (low-pass filter).

\textbf{Polynomial approximations.}  
Exact spectral filtering requires full eigendecomposition, which is impractical for large graphs. A common workaround is to approximate filters with low-order polynomials of \(\smash{\gL}\). Chebyshev expansions are particularly effective \citep{defferrard_convolutional_2016}:  
\(\smash{\widehat g_{\boldsymbol\omega}(\mLambda)=\sum_{i=0}^\rho \omega_i T_i(\tilde\mLambda)}\),  
where \(\smash{\boldsymbol\omega\in\R^{\rho+1}}\), \(\smash{\tilde\mLambda=(2/\lambda_{\max})\mLambda-\mI}\), and \(\smash{T_i}\) are Chebyshev polynomials defined recursively by \(\smash{T_0(x)=1,\;T_1(x)=x,\;T_i(x)=2xT_{i-1}(x)-T_{i-2}(x)}\). Such filters require no eigendecomposition and correspond to message passing within at most \(\rho\) hops on the graph \citep{balcilar_analyzing_2020}.

%% file: sections/3-polynomial-filters.tex
\section{Limitations of Polynomial Graph Filters}
\label{sec:bg.polynomial}

Approximating spectral filters with polynomials avoids explicit eigendecomposition, since powers of \(\gL\) act directly in the vertex domain. For example, a monomial expansion \(\smash{\widehat g_{\boldsymbol\omega}(\mLambda)=\sum_{i=0}^\rho \omega_i \mLambda^i}\) yields the spatial operator \(\smash{g_{\boldsymbol\omega}(\gL)=\sum_{i=0}^\rho \omega_i \gL^i}\). The Chebyshev basis follows an analogous reasoning (see \cref{app:proofs.poly_spatial_spectral}). While computationally attractive, such filters face inherent expressivity barriers that explain the limitations of many GNN architectures built upon them. We outline these issues from three complementary perspectives: (\textit{i}) message propagation, (\textit{ii}) function approximation, and (\textit{iii}) convergence to discontinuous filters.

\textbf{Perspective I: Propagation.}  
\citet{balcilar_analyzing_2020} show that an \(l\)-layer message-passing GNN corresponds exactly to an \(l\)-order polynomial filter. In message-passing networks, each layer typically uses a first-order update of the form \(\smash{\mH^{(l)}=(\omega_0\mI+\omega_1\gL)\mH^{(l-1)}}\), where each application of \(\gL\) aggregates information from one-hop neighbors \citep{kipf_semi-supervised_2016}. Stacking \(l\) such layers yields a polynomial filter of order \(l\), enabling information flow across at most \(l\) hops. Beyond this range, information is inaccessible, and increasing depth triggers \emph{over-squashing} \citep{alon_bottleneck_2021}, where distant signals collapse into compressed representations. Thus polynomial filters cannot achieve effective long-range propagation without prohibitive computational costs or architectural limitations.

\textbf{Perspective II: Approximation.}   
Polynomials of low degree are also weak function approximators. By Markov's inequality \citep{markov_question_1889}, the slope of a degree-\(\rho\) polynomial on \([-1,1]\) is bounded as \(\max_{x \in [-1,1]} \big|P'(x)\big| \leq \rho^2,\) which restricts the ability of the polynomial to approximate functions with sharp and steep transitions. Smooth functions can be represented with modest degree, but scaling functions and low-scale wavelets, which exhibit steep cutoffs, require prohibitively high \(\rho\). This confines polynomial filters to overly smooth spectral responses.

\textbf{Perspective III: Convergence.}  
The most fundamental limitation arises with discontinuous filters such as ideal band-pass or virtual-node mechanisms. \citet{geisler2024spatiospectral} show that polynomial approximations converge arbitrarily slowly in operator norm, regardless of degree. In other words, there exists no sequence of polynomials that can efficiently approximate such filters across graphs. This establishes a structural gap between polynomial parametrizations and the expressive, frequency-selective operators required for modeling long-range dependencies.

\begin{tcolorbox}[insightbox]
\textbf{Summary:} Polynomial filters are efficient but inherently local, spectrally smooth, and cannot represent discontinuous frequency-selective behaviors needed for long-range modeling.
\end{tcolorbox}

%% file: sections/4-method.tex
\section{Long-Range Graph Wavelet Networks}
\label{sec:method}

We propose the \emph{Long-Range Graph Wavelet Network (\modelname{})}, a parametrization of graph wavelet filters that enables efficient, principled, and expressive filtering of graph signals, specifically tailored to capture long-range dependencies. 
Our design addresses three core limitations of prior approaches: it resolves the locality bottleneck of polynomial filters, preserves the interpretability of pure wavelet propagation, and achieves linear complexity with respect to the number of edges. 
%

\subsection{Hybrid Parametrization of Wavelet Filters}
\label{sec:method.hybrid}

Effective graph filters must balance \emph{efficiency}, \emph{expressivity}, and \emph{theoretical grounding}. Polynomial filters are attractive because they can be implemented recursively with linear cost in the number of edges. However, their inherent smoothness makes them poorly suited to approximate the sharp spectral transitions characteristic of wavelets. In contrast, purely spectral filters can realize principled wavelet propagation, but computing the full eigendecomposition of the Laplacian is prohibitive.

To reconcile these trade-offs, we introduce a \emph{hybrid parametrization} that combines a polynomial backbone with a spectral correction. We recall that our goal is to parametrize wavelet filters, namely the scaling function \(\smash{\widehat{\phi}}\) and wavelet function \(\smash{\widehat{\psi}}\). At each layer \(l\), both the wavelet kernel \(\smash{\widehat{\psi}^{(l)}}\) and the scaling function \(\smash{\widehat{\phi}^{(l)}}\) are modeled as the sum of local and global contributions:
\begin{equation}
    \widehat{\psi}^{(l)}(\mLambda)
    =
    P_{\boldsymbol{\omega}_\psi^{(l)}}(\mLambda)
    +
    S_{\boldsymbol{\theta}_\psi^{(l)}}(\mLambda),
    \qquad
    \widehat{\phi}^{(l)}(\mLambda)
    =
    P_{\boldsymbol{\omega}_\phi^{(l)}}(\mLambda)
    +
    S_{\boldsymbol{\theta}_\phi^{(l)}}(\mLambda),
    \label{eq:psi_phi}
\end{equation}
where \(\smash{P_{\boldsymbol{\omega}_\bullet^{(l)}}}\) is a finite-order polynomial (spatial/local component) and \(\smash{S_{\boldsymbol{\theta}_\bullet^{(l)}}}\) is a learnable spectral (global) component. The polynomial term efficiently captures localized interactions, while the spectral term affords precise control over selected frequency ranges, crucial for long-range propagation.
\begin{lemma}
    \label{lemma:model_filtering}
    For a graph signal \(\vx \in \R^n\) and a filter kernel \(\kappa\) parametrized as in \cref{eq:psi_phi}, the filtering operation can be expressed as
    \[
        \kappa \ast \vx \;=\; \mU \!\left[P(\mLambda)+S(\mLambda)\right]\mU^\top \vx
        \;=\; P(\gL)\vx \;+\; \mU S(\mLambda)\mU^\top \vx.
    \]
\end{lemma}
This decomposition highlights the dual nature of our filtering mechanism: the polynomial term \(P(\gL)\vx\) acts directly on \(\gL\) (i.e., the vertex domain), while the spectral correction \(\smash{\mU S(\mLambda)\mU^\top \vx}\) acts on the eigenvalues (i.e., the spectral domain) and introduces frequency-selective adjustments. A proof is provided in \cref{app:proofs}.
Accordingly, at layer \(l\) we define
\begin{align}
\psi^{(l)}(\mU,\mLambda,\gL)\vx
&=
\big[P_{\boldsymbol{\omega}_\psi^{(l)}}(\gL)
\;+\;
\mU S_{\boldsymbol{\theta}_\psi^{(l)}}(\mLambda)\mU^\top\big]\vx,
\label{eq:param_psi_final}\\[2pt]
\phi^{(l)}(\mU,\mLambda,\gL)\vx
&=
\big[P_{\boldsymbol{\omega}_\phi^{(l)}}(\gL)
\;+\;
\mU S_{\boldsymbol{\theta}_\phi^{(l)}}(\mLambda)\mU^\top\big]\vx.
\label{eq:param_phi_final}
\end{align}

\textbf{Truncated spectrum.}
In practice, we compute a partial eigendecomposition \((\mU,\mLambda)=\operatorname{EVD}(\gL,k)\) of lowest \(k\) eigenpairs with \(\smash{k \ll n}\), i.e., \(\mU \in \R^{n \times k}, \mLambda \in \R^{k \times k}\). This focuses the spectral component on informative frequencies (e.g., low \(\lambda\) for long-range effects) while maintaining scalability. In practice, truncating to the lowest \(k\) eigenpairs preserves most of the signal power, since long-range interactions are dominated by low-frequency modes. The decomposition costs \(\gO(km)\), where \(m \ll n^2\) is the number of edges, is computed once at preprocessing time and reused across layers.

\textbf{Filter parametrization.}
We implement the spatial filter \(P(\gL)\) using a Chebyshev basis for stable, recursive evaluation, and the spectral filter \(S(\mLambda)\) via Gaussian smearing \citep{schutt_schnet_2017}, enabling fine-grained, band-specific control. Implementation details are given in \cref{app:impl.filter_param}.

\subsection{Principled Wavelet Propagation}
\label{sec:method.propagation}

A central feature of \modelname{} is that propagation is realized entirely through wavelet operators. Each layer applies a scaling function \(\smash{\widehat{\phi}}\) and a family of wavelet functions \(\smash{\{\widehat{\psi}_j}\}_{j=1}^J\), followed by pointwise nonlinearities and aggregation. The result is a propagation process that consists \textit{solely} of wavelet-based transformations, preserving an interpretable, wavelet-based filtering operation.

Formally, given input features \(\mH^{(l-1)}\), one \modelname{} layer computes
\begin{equation}
\mH^{(l)} =
\sigma\!\left[\phi^{(l)}(\mU,\mLambda,\gL)\,f_{\boldsymbol{\vartheta}}^{(l)}(\mH^{(l-1)})\right]
\;\oplus\;
\bigoplus_{j=1}^J
\sigma\!\left[\psi_j^{(l)}(\mU,\mLambda,\gL)\,f_{\boldsymbol{\vartheta}}^{(l)}(\mH^{(l-1)})\right],
\end{equation}
where \(\sigma\) is a nonlinearity, \(\oplus\) denotes aggregation, and \(f_{\boldsymbol{\vartheta}}^{(l)}\) is a learnable feature transformation. In practice, we incorporate \(\smash{f_{\boldsymbol{\vartheta}}^{(l)}(\mH^{(l-1)})}\) into each filter and compute, for the scaling function,
\begin{equation}
    \label{eq:layer}
    \phi^{(l)}(\mU,\mLambda,\gL)\,f_{\boldsymbol{\vartheta}}^{(l)}(\mH^{(l-1)})
    =
    \mU \!\left(S_{\boldsymbol{\theta}_\phi^{(l)}}(\mLambda) \odot \big[\mU^\top f_{\boldsymbol{\vartheta}}^{(l)}(\mH^{(l-1)})\big]\right)
    +
    P_{\boldsymbol{\omega}_\phi^{(l)}}(\gL)\,f_{\boldsymbol{\vartheta}}^{(l)}(\mH^{(l-1)}),
\end{equation}
with the wavelet functions defined analogously via \(\smash{(\boldsymbol{\omega}_\psi^{(l)},\boldsymbol{\theta}_\psi^{(l)})}\). 

We can ensure the band-pass behavior of the wavelet family, hence its wavelet theoretical validity, by enforcing the admissibility condition \(\smash{\widehat{\psi}(0)=0}\), which guarantees invertibility of the wavelet transform. In practice, strict admissibility is not always necessary and relaxing it can yield slight empirical gains. A distinctive feature of \modelname{} is that it supports both regimes: admissibility can be enforced when theoretical guarantees are required, or relaxed when flexibility and performance are prioritized. In this context, the theoretical guarantees refer to the invertibility of the transform and the preservation of band-pass behavior under the admissibility condition \(\smash{\widehat{\psi}(0)=0}\). Enforcing admissibility ensures that \modelname{} defines a valid wavelet transform with full reconstruction capability, while the relaxed variant trades strict invertibility for small empirical gains. We show how to enforce the constraint in \cref{app:impl.admissibility_condition}.


\subsection{Computational Efficiency}
\label{sec:method.computational_efficiency}
\modelname{} achieves per layer linear complexity in the number of edges, \(\gO(m)\), by combining efficient polynomial filters (\(\gO(\rho m)\)) with low-rank spectral projections (\(\gO(k d n)\)). The key insight is that partial eigendecomposition with \(k \ll n\) enables global spectral control at \(\gO(k m)\) preprocessing cost, amortized across all layers. This contrasts with purely spectral methods requiring \(\gO(n^3)\) full EVD or polynomial methods limited to \(\rho\)-hop neighborhoods. Detailed analysis is provided in \cref{app:impl.complexity}.


%% file: sections/5-results.tex
\section{Empirical Results}
\label{sec:res}

We evaluate \modelname{} on graph learning tasks requiring both long-range and short-range modeling capabilities. Our evaluation follows two axes: (\textit{i}) comparing methods based on their adherence to pure wavelet propagation, distinguishing theory-compliant from theory-relaxed approaches, and (\textit{ii}) demonstrating performance across interaction scales to validate the generality of our hybrid design.

\textbf{Experimental setup.} We conduct between 5 and 10 independent runs per experiment using different random seeds and report the mean performance and standard error of the mean. Our experiments are optimized for resource efficiency, requiring less than 10 GB of GPU memory, making them feasible on widely available hardware such as Nvidia GTX 1080Ti and 2080Ti GPUs. We use the AdamW optimizer \citep{loshchilov_decoupled_2018} with cosine annealing scheduler \citep{loshchilov_sgdr_2017} and linear warmup for stable training. Early stopping based on validation metrics prevents overfitting. All results use the \textit{independent filter parametrization} (see \cref{app:impl.filter_param}). We follow standard practice by augmenting node features with positional encodings (PEs) \citep{tonshoff_where_2023}, noting that our partial EVD provides PEs for free. Additional implementation details are provided in \cref{app:impl}.

\textbf{Wavelet methods categorization.} To fairly assess wavelet-based approaches, we distinguish between methods based on their theoretical adherence. \emph{Theory-compliant} (TC) methods satisfy admissibility constraints and propagate entirely through wavelet operations. Only SGWT \citep{hammond_wavelets_2011} and \modelname{} (with admissibility) qualify as theory-compliant. \emph{Theory-relaxed} (TR) methods either incorporate auxiliary non-wavelet components (WaveGC \citep{liu_advancing_2024}) or relax theoretical constraints (LR-GWN (without admissibility), GWNN \citep{xu_graph_2019}, DEFT \citep{bastos_learnable_2023}, ASWT-SGNN \citep{liu_aswt-sgnn_2024}).

\subsection{Long-Range Tasks}
\label{sec:res.long-range}

\textbf{Datasets.} We evaluate on \textsc{Peptides-Func} (multi-label classification of 10 functional classes) and \textsc{Peptides-Struct} (node-level regression for 11 3D structural properties) from the Long-Range Graph Benchmark \citep{dwivedi_benchmarking_2023, tonshoff_where_2023}. These tasks specifically test long-range propagation in biological structures where distant interactions determine biological functions.

\textbf{Results.} \modelname{} achieves state-of-the-art performance on both datasets, as reported in \cref{tab:results_long}. Most notably, it significantly outperforms all existing theory-compliant methods (70.52 vs 60.23 on \textsc{Peptides-Func}) while maintaining theoretical guarantees. When constraints are relaxed, it surpasses all theory-relaxed baselines. It is worth noting that even the theory-compliant version of \modelname{}, achieves comparable performance to the theory-relaxed baselines, while maintaining pure wavelet interpretability and propagation.

\textbf{Training Time.} On \textsc{Peptides-Func}, training \modelname{} takes approximately 20 seconds per epoch, totaling under 2.5 hours for full training (400 epochs). The partial EVD preprocessing costs roughly 1 minute, accounting for approximately 0.7\% of total training time, highlighting that spectral augmentation via partial EVD remains computationally practical.

\input{tables/results_single}

\subsection{Short-Range Tasks}
\label{sec:res.short-range}

\textbf{Datasets.} We evaluate on \textsc{Photo} and \textsc{Computers} from Amazon co-purchase graphs \citep{mcauley_image-based_2015, shchur_pitfalls_2018}. These datasets focus on local neighborhood interactions, validating that our long-range optimization does not compromise local modeling capabilities.

\textbf{Results.} As reported in \cref{tab:results_short}, \modelname{} achieves state-of-the-art performance on \textsc{Photo} and competitive results on \textsc{Computers}, demonstrating our proposed hybrid parametrization offers the flexibility to model both global and local propagation. Contrary to long-range tasks, relaxing admissibility constraints does not consistently improve performance on local interaction datasets.

\textbf{Training Time.} On \textsc{Photo}, training \modelname{} takes approximately 3s per epoch with total training time around 20 minutes for the full run (400 epochs). The partial EVD preprocessing with \(k = 50\) costs roughly 1.3s using sparse solvers, accounting for approximately 0.1\% of total training time.

\subsection{Ablation Study}
\label{sec:app.res.ablation}

We assess the contribution of each component in \modelname{} for long-range tasks by performing an ablation study on \textsc{Peptides-Func}.

\input{tables/component_ablation}

\textbf{Component Analysis.} The spectral component proves essential for long-range modeling (3.34\% performance drop when removed), while the polynomial component contributes primarily to computational efficiency with minimal performance impact (0.24\% drop). This validates our design rationale: spectral filtering enables long-range propagation, while polynomial filtering provides efficient local aggregation.

We report additional results in \cref{app:add_res}, and discuss related work in \cref{sec:app.related}.

%% file: tables/results_single.tex
\begin{table}[t]
    \centering
    \vspace{-5pt}
    \caption{\textbf{Benchmark results by theoretical adherence.}
    Theory-compliant methods (TC) satisfy wavelet admissibility and use pure wavelet propagation.
    Theory-relaxed methods (TR) either use additional non-wavelet modules, or relax theory constraints.
    Bold/underline represent first and second best, respectively.}
    \label{tab:results_combined}
    \begin{minipage}[t]{0.49\linewidth}
        \centering
        \subcaption{\textbf{Long-range benchmarks.}}
        \label{tab:results_long}
        \resizebox{\linewidth}{!}{%
            \begin{tabular}{lcccc}
                \toprule
                & \multicolumn{2}{c}{\textsc{Peptides-Func} (\(\uparrow\))} & \multicolumn{2}{c}{\textsc{Peptides-Struct} (\(\downarrow\))} \\
                \cmidrule(lr){2-3} \cmidrule(lr){4-5}
                Method & TC & TR & TC & TR \\
                \midrule
                SGWT \citep{hammond_wavelets_2011} & \underline{60.23 \(\pm\) 0.27} & - & \underline{25.39 \(\pm\) 0.21} & - \\
                GWNN \citep{xu_graph_2019} & - & 65.47 \(\pm\) 0.48 & - & 27.34 \(\pm\) 0.04 \\
                DEFT \citep{bastos_learnable_2023} & - & 66.95 \(\pm\) 0.63 & - & 25.06 \(\pm\) 0.13 \\
                WaveGC \citep{liu_advancing_2024} & - & \underline{69.73 \(\pm\) 0.43} & - & \underline{24.95 \(\pm\) 0.07} \\
                \midrule
                \textbf{\modelname{} (ours)} & \textbf{70.52 \(\pm\) 0.29} & \textbf{72.16 \(\pm\) 0.41} & \textbf{24.63 \(\pm\) 0.07} & \textbf{24.62 \(\pm\) 0.06} \\
                \bottomrule
            \end{tabular}
        }
    \end{minipage}\hfill
    \begin{minipage}[t]{0.49\linewidth}
        \centering
        \subcaption{\textbf{Short-range benchmarks.}}
        \label{tab:results_short}
        \resizebox{\linewidth}{!}{%
            \begin{tabular}{lcccc}
                \toprule
                & \multicolumn{2}{c}{\textsc{Photo} (\(\uparrow\))} & \multicolumn{2}{c}{\textsc{Computers} (\(\uparrow\))} \\
                \cmidrule(lr){2-3} \cmidrule(lr){4-5}
                Method & TC & TR & TC & TR \\
                \midrule
                SGWT \citep{hammond_wavelets_2011} & \underline{92.45 \(\pm\) 0.15} & - & \underline{85.19 \(\pm\) 0.22} & - \\
                GWNN \citep{xu_graph_2019} & - & 94.45 \(\pm\) 0.18 & - & 90.75 \(\pm\) 0.16 \\
                ASWT-SGNN \citep{liu_general_2025} & - & 93.80 \(\pm\) 0.12 & - & 89.40 \(\pm\) 0.19 \\
                WaveGC \citep{liu_advancing_2024} & - & \underline{95.37 \(\pm\) 0.44} & - & \textbf{92.26 \(\pm\) 0.18} \\
                \midrule
                \textbf{\modelname{} (ours)} & \textbf{95.22 \(\pm\) 0.20} & \textbf{95.69 \(\pm\) 0.23} & \textbf{91.15 \(\pm\) 0.08} & \underline{91.03 \(\pm\) 0.20} \\
                \bottomrule
            \end{tabular}
        }
    \end{minipage}
    \vspace{-8pt}
\end{table}

%% file: tables/component_ablation.tex
\begin{wraptable}{tr}{0.45\textwidth}
    \centering
    \vspace{-13pt}
    \caption{\textbf{Component ablation study on \textsc{Peptides-Func}.} We report absolute and relative drops in average precision (\(\Delta_\text{AP}\)) when removing individual components from LR-GWN.}
    \label{tab:ablation_overview}
    \resizebox{\linewidth}{!}{%
    \begin{tabular}{lcc}
        \toprule
        \textbf{Ablation} & \(\Delta_\text{AP}\) (abs) & \(\Delta_\text{AP}\) (\%) \\
        \midrule
        No Spatial Component  & \(-0.0017\)  & \(-0.24\) \\
        No Spectral Component & \(-0.0239\) & \(-3.34\) \\
        \bottomrule
    \end{tabular}
    }
    \vspace{-1em}
\end{wraptable}

%% file: sections/6-discussion.tex
\section{Discussion and Conclusion}
We revisited the limitations of wavelet GNNs constrained by polynomial approximations and introduced Long-Range Graph Wavelet Networks (\modelname{}), a hybrid design that couples polynomial aggregation with spectral corrections on truncated eigenspaces. This construction unifies local and global information flow within wavelet operators, while supporting both strict admissibility for theoretical guarantees and relaxed variants for empirical gains. 

Empirically, \modelname{} achieves state-of-the-art performance on long-range benchmarks and remains competitive on short-range tasks. More importantly, \modelname{} establishes wavelet-based filters as interpretable and theoretically grounded tools for graph learning, laying the foundation for adaptive and more expressive architectures. 
\modelname{} currently assumes static graphs and fixed truncation levels. Extending it to dynamic graphs or adapting $k$ during training remains an open avenue.

With \modelname{}, we aim to advance both the theoretical foundations and practical applications of wavelet-based long-range graph representation learning.

%% file: sections/99-acknowledgments.tex
\section*{Acknowledgments}
The authors thank Marcel Kollovieh for his input on the complexity analysis; Marten Lienen, Lukas Gosch, Yan Scholten, Tim Beyer, Alessandro Palma, and Armando Bellante for their careful feedback and proofreading; F.G. thanks Denise Cocchiarella for their helpful support and insightful discussions. This research was supported by the
German Federal Ministry of Education and Research (BMBF) through grant number 031L0289C. The authors of this work take full responsibility for its content.

%% file: sections/appendix/extended_background.tex
\section{Extended Background}
\label{app:extended_bg}

This appendix provides a more detailed overview of the mathematical foundations underlying our method. We first recall core concepts from spectral graph theory, then review wavelet analysis in both Euclidean and graph domains, and finally introduce polynomial approximations that enable scalable implementations.

\subsection{Graphs and Laplacians}
A graph \(\smash{\gG=(\mA,\mX)}\) consists of \(n\) nodes, \(m\) edges, an adjacency matrix \(\smash{\mA \in \{0,1\}^{n \times n}}\), and node features \(\smash{\mX \in \R^{n \times d}}\). The degree matrix is defined as \(\smash{\mD := \mathrm{diag}(\mA\bm{1}) \in \R^{n \times n}}\), where \(\smash{\bm{1} \in \R^n}\) denotes the all-ones vector. From this, the combinatorial Laplacian is \(\smash{\mL := \mD - \mA}\), while the symmetrically normalized Laplacian is \(\smash{\gL := \mI - \mD^{-\frac{1}{2}} \mA \mD^{-\frac{1}{2}}}\). In this work we primarily use \(\gL\), though other variants (e.g., random-walk Laplacian) are also common.  

For undirected graphs, \(\gL\) is symmetric and positive semidefinite, which guarantees an eigenvalue decomposition \(\smash{\gL = \mU \mLambda \mU^\top}\). Here \(\smash{\mU \in \R^{n \times n}}\) contains the eigenvectors as columns, forming an orthogonal basis for functions defined on the graph (\(\smash{\mU\mU^\top = \mI}\)), and \(\smash{\mLambda = \mathrm{diag}(\lambda_1,\dots,\lambda_n)}\) collects the eigenvalues. The spectrum is bounded as \(\smash{0=\lambda_1 \leq \dots \leq \lambda_n \leq 2}\). Small eigenvalues correspond to smooth, slowly varying components, while large eigenvalues capture oscillatory, high-frequency behavior \citep{chung_spectral_1997}.

\subsection{Graph Fourier Transform}
The graph Fourier transform (GFT) \citep{shuman_chebyshev_2011, sandryhaila_discrete_2013} generalizes the classical discrete Fourier transform (DFT) \citep{bracewell_fourier_2000} to signals supported on graphs. In the classical Euclidean setting, a signal is decomposed into complex exponentials, whose frequencies correspond to sinusoidal oscillations of different scales. On a graph, there is no notion of translation or frequency in the usual sense. Instead, the eigenvectors of the Laplacian play the role of Fourier modes, with the associated eigenvalues serving as the notion of frequencies.  

Given a graph signal \(\smash{\vx \in \R^n}\) (often corresponding to a column of the feature matrix \(\mX\)), the GFT is \(\smash{\widehat{\vx} = \mU^\top \vx}\), which projects the signal onto the eigenbasis of \(\gL\). The inverse GFT reconstructs the signal as \(\smash{\vx = \mU \widehat{\vx}}\). Filtering with a spectral kernel \(\smash{g:\R\to\R}\) amounts to rescaling each Fourier coefficient according to the eigenvalue, yielding
\[
\smash{g \ast_{\gG} \vx = \mU \, \widehat g(\mLambda) \, \mU^\top \vx},
\]
where \(\smash{\widehat g(\mLambda) = \mathrm{diag}(\widehat g(\lambda_1),\dots,\widehat g(\lambda_n))}\), and \(\widehat g = \mU^\top g\). In this sense, spectral filters are functions of the Laplacian spectrum, directly analogous to frequency filters in classical Fourier analysis.

\subsection{Wavelet Analysis}
Wavelet theory extends Fourier analysis by enabling localized, multiresolution representations of signals \citep{mallat_multiresolution_1989, mallat_wavelet_2009}. Whereas Fourier bases are global and capture only frequency information, wavelets provide both spectral and spatial localization by combining scaling and translation operations.  

The continuous wavelet transform (CWT) of a function \(\smash{f:\R\to\R}\) at scale \(\smash{s>0}\) and location \(\smash{l\in\R}\) is defined as
\[
W_f(s,l) = \int_\R f(t)\,\psi^*_{s,l}(t)\,dt,
\]
where the wavelet family is generated from a mother wavelet \(\smash{\psi}\) by scaling and translation, \(\smash{\psi_{s,l}(t) = \frac{1}{\sqrt{s}}\psi\!\left(\frac{t-l}{s}\right)}\), and \(^*\) denotes complex conjugation. Exact reconstruction requires the mother wavelet to satisfy the admissibility condition, ensuring that no frequency component is lost.

\begin{proposition}[Wavelet Admissibility]
\label{prop:admissibility_condition}
A wavelet \(\smash{\psi}\) with Fourier transform \(\smash{\widehat{\psi}}\) is admissible if
\[
C_\psi = \int_0^\infty \frac{|\widehat{\psi}(\zeta)|^2}{\zeta}\,d\zeta < \infty,
\]
which holds when \(\smash{\widehat{\psi}(0)=0}\) and \(\smash{\lim_{\zeta\to\infty}\widehat{\psi}(\zeta)=0}\).
\end{proposition}

This condition ensures that \(\smash{\widehat{\psi}}\) behaves as a band-pass filter, emphasizing intermediate frequency bands while attenuating very low and high frequencies. To retain the low-frequency components, wavelet systems are typically complemented by a scaling function \(\smash{\phi}\) with Fourier transform \(\smash{\widehat{\phi}(0)>0}\), which acts as a low-pass filter. Together, \(\phi\) and the wavelet family provide a multiresolution decomposition of signals.

\subsection{Wavelets on Graphs}
\citet{hammond_wavelets_2011} extend wavelet analysis to graphs by defining filters in the Laplacian spectral domain. Let \(\smash{\gL=\mU\mLambda\mU^\top}\) be the eigendecomposition of the (normalized) Laplacian, and let \(\smash{\widehat{\psi}}\) and \(\smash{\widehat{\phi}}\) denote a band-pass wavelet kernel and a low-pass scaling kernel, respectively (both applied elementwise to eigenvalues).

The \emph{wavelet operator} at scale \(\smash{s>0}\) and the \emph{scaling operator} are
\[
\smash{\Psi_s := \mU\,\widehat{\psi}(s\mLambda)\,\mU^\top}, 
\qquad
\smash{\Phi := \mU\,\widehat{\phi}(\mLambda)\,\mU^\top}.
\]
Given a graph signal \(\smash{\vx\in\R^n}\), the wavelet coefficients at scale \(\smash{s}\) are \(\smash{W_\vx(s)=\Psi_s\vx}\), while the scaling coefficients are \(\smash{H_\vx=\Phi\vx}\). This parallels the classical construction: wavelets capture localized variations, whereas the scaling function captures smooth, global structure. A multiscale analysis is obtained by a family of dilations \(\smash{\{s_j\}}\), i.e., \(\smash{\{\widehat{\psi}(s_j\cdot)\}_j}\). On graphs, admissibility is typically enforced by \(\smash{\widehat{\psi}(0)=0}\), with low frequencies covered by \(\smash{\widehat{\phi}}\) satisfying \(\smash{\widehat{\phi}(0)>0}\).

\subsection{Chebyshev Polynomial Approximations}
Although spectral constructions are mathematically elegant, direct computation is expensive due to the eigenvalue decomposition, which scales cubically in the number of nodes. To make spectral filters scalable, they are commonly approximated with low-order polynomials in the Laplacian, which can be applied directly in the vertex domain.  

A widely used choice is the Chebyshev polynomial basis \citep{hammond_wavelets_2011, defferrard_convolutional_2016}. For a filter \(\smash{\widehat g}\), one approximates
\[
\widehat g_{\boldsymbol\omega}(\mLambda) = \sum_{i=0}^\rho \omega_i T_i(\tilde{\mLambda}),
\]
where \(\smash{\boldsymbol\omega=[\omega_0,\dots,\omega_\rho]}\) are learnable coefficients, \(\smash{\rho}\) is the polynomial order, and \(\smash{\tilde{\mLambda} = \tfrac{2}{\lambda_{\max}}\mLambda - \mI}\) rescales the spectrum to \([-1,1]\), the domain of the Chebyshev basis.

\begin{definition}[Chebyshev Polynomials of the First Kind]
The Chebyshev polynomials \(\smash{T_i}\) are defined recursively by
\[
T_0(x)=1,\quad T_1(x)=x,\quad T_i(x)=2xT_{i-1}(x)-T_{i-2}(x)\quad (i\geq 2).
\]
\end{definition}

Chebyshev expansions are particularly effective because they minimize the maximum approximation error on \([-1,1]\) (in the minimax sense), and can be evaluated efficiently via recurrence. In practice, this enables fast, localized, and scalable approximations of spectral filters, including graph wavelets, without explicit eigendecomposition.

%% file: sections/appendix/polynomial_filter_limitations.tex
\section{Extended Discussion on Polynomial Filter Limitations}
\label{app:poly_limits}

Polynomial filters have become the standard choice for approximating spectral graph operators because they can be applied without eigendecomposition. In this section, we expand on the three perspectives introduced in \cref{sec:bg.polynomial}, providing formal statements and illustrative examples.

\subsection{Propagation and Hop-Limits}
The interpretation of polynomial filters as message-passing operators is immediate from the fact that \(\gL^k\) encodes walks of length \(k\) \citep{balcilar_analyzing_2020}. A degree-\(\rho\) polynomial filter can therefore only aggregate information from nodes within \(\rho\) hops of each target node. This locality is beneficial for efficiency, but also imposes a hard ceiling on the effective receptive field. Deep message-passing GNNs, which correspond to higher-degree polynomials, in principle extend this range but suffer from \emph{over-squashing} \citep{alon_bottleneck_2021}: as the neighborhood grows exponentially with hop count, long-range signals are compressed into fixed-dimensional node embeddings, leading to severe information loss. Thus, regardless of how coefficients are chosen, polynomial filters cannot escape their fundamentally local nature.

\input{figures/wavelet_approximation}

\subsection{Function Approximation and Smoothness}
From a spectral perspective, the limitations of polynomial filters manifest in their inability to approximate functions with steep slopes. This is formally captured by Markov's inequality:

\begin{theorem}[Markov's Inequality \citep{markov_question_1889}]
Let \(P\) be a polynomial of degree at most \(\rho\) with \(|P(x)|\leq 1\) for all \(x \in [-1,1]\). Then
\[
\max_{x \in [-1,1]} \left|\frac{dP(x)}{dx}\right| \leq \rho^2.
\]
\end{theorem}

The theorem shows that the slope of any polynomial is globally bounded, preventing sharp transitions. Consequently, low-order polynomials can approximate smooth filters, but require very high degree to approximate narrow band-pass filters or scaling functions with steep low-frequency cutoffs. 
In \cref{fig:true_vs_approx}, we visualize this phenomenon: wavelets at medium scales are well approximated even with low-order expansions, while scaling functions and fine-scale wavelets are captured poorly unless the degree is significantly increased.

\subsection{Convergence to Discontinuous Filters}
A more severe limitation emerges when considering discontinuous spectral filters, such as ideal low-pass or band-pass functions. While continuous functions can, in principle, be approximated uniformly by polynomials (by the Weierstrass theorem), discontinuous functions break this guarantee. Recent results make this precise in the context of graph filters:

\begin{theorem}[Slow Polynomial Convergence \citep{geisler2024spatiospectral}]
Let \(\widehat g\) be a discontinuous spectral filter. For any polynomial sequence \((g_{\gamma_\rho})_{\rho}\), there exists a graph sequence \((\gG_\rho)_{\rho}\) such that
\[
\nexists \alpha > 0:\;
\sup_{0\neq \mX \in \R^{|\gG_\rho|\times d}}
\frac{\|(g_{\gamma_\rho}-g)\ast_{\gG_\rho}\mX\|_F}{\|\mX\|_F}
= \mathcal{O}(\rho^{-\alpha}).
\]
\end{theorem}

This result formalizes the intuition that discontinuous filters cannot be efficiently represented by polynomials: convergence is arbitrarily slow in operator norm, regardless of polynomial degree. This barrier persists across graphs of increasing size, implying that polynomial parametrizations are structurally mismatched with frequency-selective or discontinuous behaviors.

\subsection{Takeaways}
Together, these results reveal the inherent trade-off in polynomial filters: they offer computational simplicity but at the cost of locality, smoothness restrictions, and poor convergence on discontinuous operators. These limitations highlight the need for alternative parametrizations—such as wavelet-based designs—that can preserve efficiency while enabling non-local, multi-scale, and spectrally expressive behavior.

%% file: figures/wavelet_approximation.tex
\begin{figure}[t]
    \centering
    \includegraphics[width=0.5\linewidth]{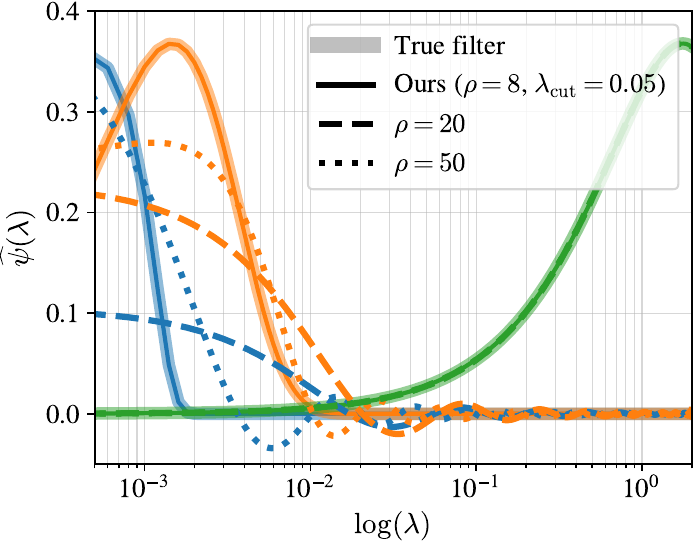}
    \caption{
    Different polynomial approx.\ vary in their ability to model low-frequency wavelets. Both low-order (\(\rho = 20\)) and high-order (\(\rho = 50\)) variants struggle with sharp transitions. Our method (\(\rho = 8, \lambda_\text{cut}=0.05\)) captures both smooth and sharp irregularities.}
    \label{fig:true_vs_approx}
\end{figure}

%% file: sections/appendix/related_work.tex
\section{Related Work}
\label{sec:app.related}

\textbf{Wavelets on Graphs.}
The Spectral Graph Wavelet Transform (SGWT) by \citet{hammond_wavelets_2011} introduced cubic spline wavelets and an exponential scaling function at fixed scales, ensuring admissibility but lacking learnable parameters. SGWT approximates filters using finite-order Chebyshev polynomials.
Graph Wavelet Neural Networks (GWNN) \citep{xu_graph_2019} extend SGWT with fixed exponential wavelets, i.e., \(g(s\lambda_i) = e^{s\lambda_i}\). While building on wavelet theory, this construction does not enforce the admissibility condition, since \(g(0)=1\).
DEFT \citep{bastos_learnable_2023} follows SGWT and parametrizes wavelet filters using free-parameters Chebyshev polynomials via a combination of GNNs and MLPs. The coefficients, however, are not bound and, by construction, do not enforce the admissibility condition. 
ASWT-SGNN \citep{liu_aswt-sgnn_2024} approximates wavelet filter coefficients using Jackson–Chebyshev polynomials. Similarly to the discussion above about GWNN and DEFT, the coefficients are not bound to satisfy \(g(0)=0\).
AGT \citep{cho_neurodegenerative_2024} extends GWNN, and learns node-wise scales, either exactly or via approximation. As with the methods discussed so far, this approach does not enforce the admissibility condition, and thus does not guarantee that the resulting filters behave as band-pass wavelets.
WaveGC \citep{liu_general_2025} satisfies the admissibility condition by parametrizing the scaling function with odd-terms of a Chebyshev polynomial, and the wavelet function with even-terms. However, the separation of basis terms limits functional expressiveness, and the use of non-wavelet-interpretable message passing diminishes the benefit of using wavelets.

\textbf{Combining Spatial and Spectral Filters for Long-Range Interactions.}
Hybrid GNNs combine local spatial and global spectral information to enhance long-range modeling. \citet{stachenfeld_graph_2020} propose a hybrid message-passing framework that sacrifices permutation equivariance, while \citet{liao_lanczosnet_2018} incorporate spectral information via the Lanczos algorithm.
More recently, \citet{geisler2024spatiospectral} introduced a spatial-spectral parametrization to improve efficiency. Beyond the explicit combination of spatial and spectral filters, graph rewiring \citep{gasteiger_diffusion_2019, arnaiz-rodriguez_diffwire_2022} enhance connectivity and mitigate over-squashing by modifying connectivity.
Graph transformers (GTs) \citep{min_transformer_2022, dwivedi_graph_2021, geisler_transformers_2023} use global attention mechanisms to enable interactions between distant nodes. GTs often rely on spectral positional encodings paired with message passing \citep{rampasek_recipe_2023, geisler_transformers_2023}. \citet{rampasek_recipe_2023} introduce a framework for general, powerful and scalable graph transformers (GPS), which allow for the definition of a broad family of architectures that combine local message passing with global attention mechanisms. Similarly to these models, LR-GWN integrates local and global components. In our case, local aggregation is achieved through polynomial filters, while global propagation is handled via a spectral wavelet filter. However, the core methodology and motivation differ. LR-GWN is rooted in wavelet theory, with the aim of modeling low-pass and band-pass filters in a principled and interpretable manner. Graph transformers, by contrast, cover a broad family of architectures with varying attention mechanisms and are generally not designed with the same frequency-selective properties in mind.

Our mention of graph transformers is intended purely for conceptual contrast; they were not included in experimental comparisons, as our focus is restricted to wavelet-based operators.

%% file: sections/appendix/implementation.tex
\section{Implementation Details}
\label{app:impl}

\subsection{Filter Parametrization}
\label{app:impl.filter_param}

\textbf{Spatial Filter.}
We model the spatial filter \(P_{\boldsymbol{\omega}^{(l)}}: [0, \lambda_\text{max}]^k \to \R^k\) using a finite-order Chebyshev polynomial expansion with learnable coefficients \(\boldsymbol{\omega}^{(l)} = [\omega_0^{(l)}, \dots, \omega_{\rho}^{(l)}] \in \R^{\rho+1}\):
\begin{equation}
    \label{eq:spatial_filter}
    P_{\boldsymbol{\omega}^{(l)}}(\gL) = \sum_{i=0}^{\rho} \omega_i^{(l)} T_i(\gL),
\end{equation}
where \(T_i(\cdot)\) is the \(i\)-th Chebyshev polynomial. This formulation enables spatially localized filtering with computational efficiency. Alternative polynomial bases, such as Bernstein \citep{he_bernnet_2021} or Jacobi polynomials \citep{wang_how_2022}, could also be employed.

\textbf{Spectral Filter.}
The spectral component \(S_{\boldsymbol{\theta}^{(l)}}: [0, \lambda_\text{max}]^k \to \R^k\) operates directly on the eigenvalues of \(\gL\), offering fine-grained frequency control. It is defined as
\begin{equation}
    \label{eq:spectral_filter}
    S_{\boldsymbol{\theta}^{(l)}}(\lambda) = \operatorname{GaussianSmearing}(\lambda) \mW_{\theta^{(l)}},
\end{equation}
where \(\mW_{\theta^{(l)}} \in \R^{z \times d^{(l)}}\) are learnable weights, and \(d^{(l)}\) is the output dimensionality of layer \(l\). The basis \(\operatorname{GaussianSmearing}(\lambda): [0,2]^k \to \R^{k \times z}\) projects eigenvalues onto \(z\) Gaussian radial functions, introducing a frequency cutoff \(\lambda_\text{cut} < \lambda_\text{max}\) \citep{schutt_schnet_2017}. We follow \citet{geisler2024spatiospectral} and apply a frequency-domain windowing function to regularize the spectral response. Element-wise, the filter is defined as \(\smash{S_{\boldsymbol{\theta}^{(l)}}(\mLambda)_{u,v} := S_v(\lambda_u; \boldsymbol{\theta}^{(l)})}\).
Alternative spectral parametrizations, such as SpecFormer \citep{bo_specformer_2023}, could be used depending on application-specific needs.

\subsubsection{Shared and Independent Filter Configurations}
\label{app:impl.shared-independent}

We define two configurations for our wavelet-based graph neural network: the \textit{shared filter} and the \textit{independent filter} approaches. In the shared filter configuration, we parametrize a single mother wavelet and generate a family of self-similar wavelet functions by applying predefined or learnable scales. Conversely, in the independent filter approach, each wavelet function is parametrized independently, allowing greater flexibility and expressivity.

\textbf{Shared Filter.} 
In this configuration, the \(j\)-th wavelet filter is constructed as
\begin{equation}
    \label{eq:single_filter_wavelet}
    \psi\left(\mU, \mLambda, \gL; s_j\right) = \mU S_{\theta_\psi}(s_j \mLambda) \mU^\top + P_{\omega_\psi}(s_j \gL),
\end{equation}
where a single function \(\psi\) is parametrized following \cref{eq:param_phi_final} and then scaled using predefined or learnable scales (\cref{app:impl.scale}). This method is computationally efficient, as it requires only a single set of parameters \(\theta_\psi\) and \(\omega_\psi\), reducing the overall complexity of the model.

\textbf{Independent Filters.} 
To increase model expressivity, we introduce an alternative approach in which each wavelet function is parametrized independently instead of being derived from a single mother wavelet. This configuration allows each wavelet to adjust its spectral and polynomial components separately, adapting to different levels of smoothness or sharpness. 
Larger-scale wavelets tend to be smoother and benefit more from polynomial components, while smaller-scale wavelets are sharper and require a stronger contribution from the spectral component. 
In the independent filters configuration, the \(j\)-th wavelet filter is given by
\begin{equation}
    \label{eq:multi_filter_wavelet}
    \psi_j\left(\mU, \mLambda, \gL\right) = \mU S_{\theta_{\psi_j}}(\mLambda) \mU^\top + P_{\omega_{\psi_j}}(\gL),
\end{equation}
where each \(\psi_j\) has its own set of learnable spectral parameters \(\theta_{\psi_j}\) and polynomial parameters \(\omega_{\psi_j}\). 

\subsubsection{Scale Parametrization.}
\label{app:impl.scale}
Wavelet-based graph filters rely on a set of scales \(\{s_j\}_{j=1}^J\) to capture multi-resolution information. Instead of using fixed scales as in \citet{hammond_wavelets_2011}, we introduce a learnable interpolation scheme in the shared filter configuration (\cref{app:impl.shared-independent}). The scale bounds are defined as
\(\log s_{\min} = \log (t_{\text{L}}/\lambda_\text{max})\) and \(\log s_{\max} = \log (t_{\text{U}} \lambda_{\text{LP}}/\lambda_\text{max})\),
where \(t_{\text{L}}\) and \(t_{\text{U}}\) are learnable lower and upper limits, and \(\lambda_{\text{LP}}\) emphasizes low-frequency components. Assuming the maximum eigenvalue of the graph to be \(\lambda_{\max} = 2\) as it is usually the case in real-world graphs, we have \(\log s_{\min} = \log t_{\text{L}}, \log s_{\max} = \log t_{\text{U}} \lambda_{\text{LP}}\). The intermediate scales are uniformly spaced in log-space,
\begin{equation}
\label{eq:scales}
s_i = \exp\left( \log s_{\max} + \frac{i (\log s_{\min} - \log s_{\max})}{N-1} \right).
\end{equation}
To ensure positive scales, we apply the \(\operatorname{softplus}\) function.

\subsection{Modeling Practices}

\subsubsection{Initialization}
\label{app:impl.initialization}

The wavelet filter parameters are initialized using standard Xavier initialization \citep{glorot_understanding_2010} for neural network weights and by setting \(\omega_0 = 1\) and \(\omega_i = 0\) for all \(i = 1, \dots, \rho\) in the Chebyshev polynomials. While the spectral and spatial components could theoretically be designed to match a predefined wavelet, such as the Mexican hat wavelet, we observed no significant improvements from this approach. Consequently, we adopt conventional parameter initialization without enforcing wavelet-based constraints in our experiments.

\subsubsection{Admissibility Condition}
\label{app:impl.admissibility_condition}

To ensure the theoretical validity of our graph wavelet filter, we must guarantee the invertibility of the wavelet transform. This requires satisfying the admissibility condition at zero frequency, as stated in \cref{prop:admissibility_condition}. Since the graph spectrum is bounded (i.e., \(\lambda \in [0, \lambda_{\max}]\) with \(\lambda_{\max} \leq 2\)), we only need to enforce the condition \(\smash{\widehat{\psi}(0) = 0}\), while disregarding \(\smash{\lim_{\lambda \to \infty} \widehat{\psi}(\lambda) = 0}\).

Enforcing the admissibility condition is not always a necessity. Rather, it is a theoretical property that ensures the wavelet behaves as a band-pass filter. In practice, many applications do not strictly require this condition, and relaxing it can lead to slight empirical gains in some cases.
However, with LR-GWN, we aim to provide a theoretically grounded framework that can allow to enforce the admissibility condition if needed, while also permitting for flexibility in applications where this condition is not strictly required. 

\paragraph{Primary Approach.}
In practice, we can enforce this constraint by subtracting the zero-frequency response from the spectral and polynomial filters:
\begin{equation}
    \widetilde{P}_\omega(\lambda) = P_\omega(\lambda) - P_\omega(0), \quad
    \widetilde{S}_\theta(\lambda) = S_\theta(\lambda) - S_\theta(0).
\end{equation}
This transformation ensures \(\widehat{\psi}(0) =  \widetilde{P}_\omega(0) + \widetilde{S}_\theta(0) = 0\) by construction.
Thus, the wavelet filter in the spectral domain becomes
\begin{equation}
    \widehat{\psi}(\mLambda) = 
    \widetilde{P}_\omega(\mLambda),
    + 
    \widetilde{S}_\theta(\mLambda) 
\end{equation}
and the filtering operation in the vertex domain is given by
\begin{equation}
    \psi\left(\mU, \mLambda, \gL\right) = 
    \widetilde{P}_\omega(\gL),
    + 
    \mU \widetilde{S}_\theta(\mLambda) \mU^\top 
\end{equation}
where both \(\widetilde{P}_\omega(\gL)\) and \(\widetilde{S}_\theta(\mLambda)\) are zero-frequency corrected versions of their respective filters.

\paragraph{Alternative Approach.}
For the wavelet filter \(\psi(\mLambda) = S_{\theta}(\mLambda) + P_{\omega}(\mLambda)\), where \(P_{\omega}(\mLambda)\) is expressed as a Chebyshev polynomial, the admissibility condition at zero frequency is satisfied if
\[
\omega_0 = \sum_{i=1}^\rho (-1)^{i+1} \omega_{i} - S_{\theta}(0).
\]
We show the steps that lead to the above formulation. The wavelet filter \(\psi(\lambda)\) consists of two components:
\[
\psi(\lambda) = S_{\theta}(\lambda) + P_{\omega}(\lambda),
\]
where \(S_{\theta}(\lambda)\) and \(P_{\omega}(\lambda)\) are the spectral and polynomial filter responses, respectively, at frequency \(\lambda\). The admissibility condition requires that \(\psi(0) = 0\), i.e., 
\[
S_{\theta}(0) + P_{\omega}(0) = 0.
\]
For the admissibility condition to hold, we either enforce both components to be zero (\(S_{\theta}(0) = P_{\omega}(0) = 0\)) or allow \(P_{\omega}(0) = -S_{\theta}(0)\).
We express \(P_{\omega}(\lambda)\) as a Chebyshev polynomial. After rescaling the input from \(\lambda \in [0, 2]\) to \(\widetilde{\lambda} \in [-1, 1]\), the condition \(P_{\omega}(0) = \widetilde{P}_\omega(-1)\) leads to the equation
\[
\omega_0 = \sum_{i=1}^\rho (-1)^{i+1} \omega_i - S_{\theta}(0),
\]
which satisfies the admissibility condition at zero frequency.

However, this approach is more challenging to implement, as it requires manually updating the learnable parameters. While this is feasible, we found that the primary approach worked better overall. It is more flexible and streamlined, making it easier to integrate into our model while still satisfying the necessary conditions.

\input{figures/learned_filters}

\subsubsection{Wavelet Residuals}
We introduce residual connections within the wavelet filter (\textit{wavelet residuals}) to improve model generalization. We write the forward step of our model with included wavelet residual connections as
\begin{equation}
    \hat{\mH}_{\psi,j}^{(l-1)} = \sigma\left[ \left(\psi^{(l)}\left(\mU, \mLambda, \gL\right) + \mI \right) \hat{\mH}^{(l-1)}\right] = \sigma\left[{\psi_\mI}^{(l)}\left(\mU, \mLambda, \gL\right) \hat{\mH}^{(l-1)}\right].
\end{equation}
In the graph frequency domain, this corresponds to the graph wavelet filter \(\smash{\widehat{\psi}_\mI(\mLambda) = S_\theta(\mLambda) + P_\omega(\mLambda) + \mI}\).
While including wavelet residual connections brings desirable properties in terms of gradient flow, preserving the theoretical properties of the wavelet transform remains important to theoretically guarantee the existence of the inverse wavelet transform.
Similarly to how we enforced the admissibility condition in \cref{app:impl.admissibility_condition}, we can now let the spectral and polynomial filters transform as
\(S_\theta(\lambda) \to S_\theta(\lambda) - S_\theta(0) - 1/2\), and \(P_\omega(\lambda) \to P_\omega(\lambda) - P_\omega(0) - 1/2\).
In fact, this allows us to write \(\smash{\widehat{\psi}_\mI(\mLambda) = S_\theta(\mLambda) - S_\theta(\boldsymbol{0}) - \mI/2 + P_\omega(\mLambda) - P_\omega(\boldsymbol{0}) - \mI/2 + \mI}\).
The wavelet filter then becomes \(\smash{\psi_\mI\left(\mU, \mLambda, \gL\right) = U\widetilde{S}_\theta(\mLambda)U^\top + \widetilde{P}_\omega(\gL) + \mI}\),
where \(\smash{\widetilde{S}_\theta(\mLambda) = S_\theta(\mLambda) - S_\theta(\boldsymbol{0}) - \mI/2}\) and \(\smash{\widetilde{P}_\omega(\gL) = P_\omega(\gL) - P_\omega(\boldsymbol{0}) - \mI/2}\), thus ensuring the condition \(\smash{\widehat{\psi}(0) = 0}\) by construction.

\paragraph{Residual Connections.}
While the inclusion of wavelet residuals theoretically ensures the preservation of the admissibility condition, in practice, we observe that normal residual connections, where the residual is added after the filtering operation, further improve the model's generalization. Therefore, we opt to use standard residual connections.

\subsection{Computational Complexity.}
\label{app:impl.complexity}

The LR-GWN layer integrates multiple structural components whose computational costs can be precisely characterized. Let \(n\) and \(m\) denote the number of nodes and edges in the input graph, and let \(d\) represent the feature dimensionality, assumed constant across all layers of the network. We denote by \(L\) the depth of the MLP \(\smash{f_\vartheta^{(l)}}\), by \(\rho\) the polynomial order used in the Chebyshev approximation, by \(k\) the number of retained eigenvectors for the spectral component, and by \(J+1\) the total number of filters, including the scaling function.

Each layer begins with a node-wise transformation via an \(L\)-layer MLP of hidden size \(d\), shared across both filtering components. This projection incurs a cost of \(\smash{\gO(L d^2 n)}\). The polynomial component uses recursive Chebyshev polynomials over the sparse Laplacian. Recall that multiplication between an \(n \times n\) sparse matrix with \(m\) non-zero entries, and a dense \(n \times d\) matrix has cost \(\gO(md)\). For a \(\rho\)-order polynomial, and across all filters, this results in a total cost of \(\smash{\gO((J+1)\rho md)}\). The spectral component computes a projection via \(\mU\) and back, for each filter. This involves two matrix multiplications of cost \(\gO(k d n)\) and an element-wise scaling of cost \(\gO(k d)\), which is asymptotically negligible. Across filters, the spectral component contributes a total cost of \(\gO((J+1) k d n)\).

Crucially, the partial eigendecomposition required for the spectral component is computed once per graph at cost \(\gO(k m)\) as a preprocessing step, and can be amortized across the network.

In summary, the per-layer complexity is \(\gO\left(L d^2 n + (J+1)\rho d m + (J+1) k d n\right)\).\
Since \(J\), \(L\), \(d\), \(\rho\), and \(k\) are typically small constants in practice, the dominant terms grow with the number of nodes \(n\), and edges \(m\). The expression therefore simplifies to \(\gO(n + m)\). For sparse graphs we can assume constant node degree \(\text{deg}\), hence \(m = \text{deg} \cdot n = \Omega(n)\), yielding an asymptotical complexity of \(\gO(m)\).
This decomposition reveals the efficiency of LR-GWN: despite leveraging both spatial and spectral graph filters, the model remains scalable to large, sparse graphs.

%% file: figures/learned_filters.tex
\begin{figure*}[!t]
    \centering
    \includegraphics[width=\textwidth]{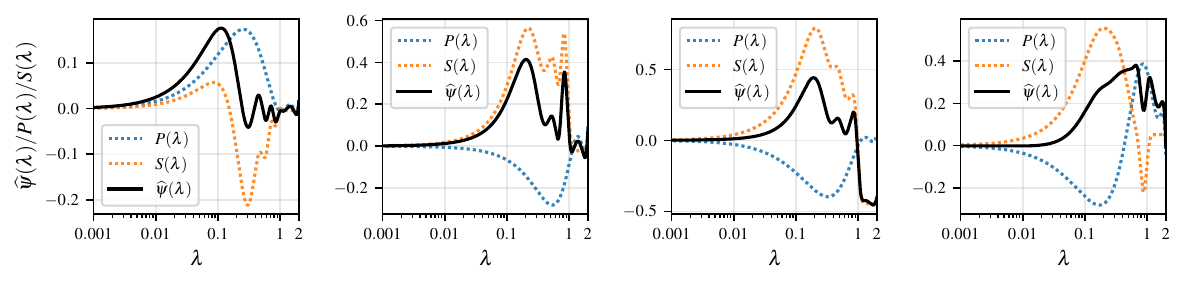}
    \caption{Examples of filters learned by our model during training. The polynomial component \(P(\lambda)\) captures the smooth part of the filter, while the spectral component \(S(\lambda)\) introduces flexibility, enabling sharper variations. Our filters inherently satisfy the admissibility condition \(\smash{\widehat{\psi}(0) = 0}\).}
    \label{fig:filters}
\end{figure*}

%% file: sections/appendix/additional_results.tex
\section{Additional results}
\label{app:add_res}

We report here additional results to those presented in \cref{sec:res}.

\subsection{Ablation Study}
\label{app:add_res.ablation}

\textbf{Admissibility Trade-off.} The performance difference between theory-compliant (70.52) and theory-relaxed (72.16) variants on \textsc{Peptides-Func} demonstrates \modelname{}'s flexibility in balancing theoretical guarantees with empirical performance, allowing adaptation to application-specific requirements.

\textbf{Considerations on \(k\) and \(\rho\).} We conducted preliminary ablation studies on both the number of retained eigenvalues \(k\) and polynomial order \(\rho\). While performance varies with both hyperparameters, no consistent pattern emerges across datasets. For smaller graphs like \textsc{Peptides-Func}, we retain a large number of eigenvalues (\(k = 150\)) while keeping computational costs manageable, exploring the full expressive capacity. On larger graphs, smaller \(k\) values typically suffice to balance efficiency and performance. Similarly, optimal \(\rho\) is dataset-specific, but the model remains robust within reasonable ranges (typically \(\rho = 3\)-15).

%% file: sections/appendix/proofs.tex
\section{Theoretical Results}
\label{app:proofs}

\subsection{Polynomial filters in spectral and spatial domain}
\label{app:proofs.poly_spatial_spectral}

For a graph with Laplacian \(\gL = \mU\mLambda \mU^\top\) and a polynomial filter of degree \(\rho\), the spectral filtering operation \(\mU \widehat{g}_\omega(\mLambda) \mU^\top x\) is equivalent to the spatial operation \(g_\omega(\gL) x\), where \(\widehat{g}_\omega(\mLambda) = \sum_{i=0}^\rho \omega_i \mLambda^i\).

\begin{proof}
    We prove this by showing that powers of the Laplacian in the spatial domain correspond to powers of eigenvalues in the spectral domain.
    Starting with the spectral filtering operation:
    \begin{align}
        \mU g_\omega(\mLambda) \mU^\top x &= \mU \left(\sum_{i=0}^\rho \omega_i \mLambda^i\right) \mU^\top x \\
                                          &= \sum_{i=0}^\rho \omega_i \mU \mLambda^i \mU^\top x
    \end{align}
    Now we use the key insight that for the eigendecomposition \(\gL = \mU\mLambda \mU^\top\), we have:
    \[\mU \mLambda^i \mU^\top = (\mU\mLambda \mU^\top)^i = \gL^i.\]
    
    This holds by induction:
    \begin{itemize}
        \item Base case \(i = 1\): \(\mU\mLambda \mU^\top = \gL\) by definition
        \item Inductive step: Assume \(\mU \mLambda^k \mU^\top = \gL^k\). Then:
        \begin{align}
            \mU \mLambda^{k+1} \mU^\top &= \mU \mLambda^k \mLambda \mU^\top\\
            &= \mU \mLambda^k (\mU^\top \mU) \mLambda \mU^\top \quad \text{(since } \mU^\top \mU = \mI\text{)}\\
            &= (\mU \mLambda^k \mU^\top)(\mU \mLambda \mU^\top)\\
            &= \gL^k \gL = \gL^{k+1}
        \end{align}
    \end{itemize}
    
    Therefore:
    \[\mU g_\omega(\mLambda) \mU^\top x = \sum_{i=0}^\rho \omega_i \gL^i x = g_\omega(\gL) x\]
    
    This equivalence explains why polynomial spectral filters can be implemented efficiently without eigendecomposition—they correspond exactly to polynomial operations on the Laplacian matrix itself.
\end{proof}

\textbf{Remark.} This result extends immediately to any polynomial basis (Chebyshev, Bernstein, etc.) since they are all linear combinations of monomials.

\subsection{Proof of \texorpdfstring{\cref{lemma:model_filtering}}{Lemma \ref{lemma:model_filtering}}}
\label{app:proofs.lemma_model_filtering}
\begin{proof}
    We begin by expanding the left-hand side of the expression. Distributing the product over the sum gives
    \begin{equation}
       \label{eq:expanded_sum}
       \mU S(\mLambda) \mU^\top + \mU P(\mLambda) \mU^\top.
    \end{equation}
    Next, we focus on the term \( P(\mLambda) \), which is parameterized as a polynomial function of the eigenvalues of the graph Laplacian \(\mLambda\). Specifically, \( P(\mLambda) \) can be written as
    \begin{equation}
       \label{eq:polynomial_expansion}
       P(\mLambda) = \sum_{i=0}^{\rho} \gamma_i \mLambda^i,
    \end{equation}
    where \( \gamma_i \) are learnable coefficients and \( \rho \) is the polynomial degree.
    Substituting this expansion for \( P(\mLambda) \) into the expression, we get:
    \begin{align}
       \mU P(\mLambda) \mU^\top &= \mU \left( \sum_{i=0}^{\rho} \gamma_i \mLambda^i \right) \mU^\top \label{eq:substituted_polynomial} \\
                                &= \sum_{i=0}^{\rho} \gamma_i \mU \mLambda^i \mU^\top \label{eq:distributed_sum}
    \end{align}
    The key insight is that for the eigendecomposition \(\gL = \mU \mLambda \mU^\top\), we have:
    \begin{equation}
       \label{eq:power_equivalence}
       \mU \mLambda^i \mU^\top = \gL^i.
    \end{equation}
    This holds by induction as shown in \cref{app:proofs.poly_spatial_spectral}.
    Substituting \cref{eq:power_equivalence} into \cref{eq:distributed_sum}, we obtain
    \begin{equation}
       \label{eq:spatial_polynomial}
       \mU P(\mLambda) \mU^\top = \sum_{i=0}^{\rho} \gamma_i \gL^i = P(\gL).
    \end{equation}
    Finally, combining the two terms from \cref{eq:expanded_sum}, we get the desired decomposition
    \begin{equation}
       \label{eq:final_decomposition}
       \mU \left[ S(\mLambda) + P(\mLambda)\right] \mU^\top = \mU S(\mLambda) \mU^\top + P(\gL).
    \end{equation}
    This completes the proof.
\end{proof}

%% file: lrgwn.bbl
\begin{thebibliography}{48}
\providecommand{\natexlab}[1]{#1}

\bibitem[Alon \& Yahav(2021)Alon and Yahav]{alon_bottleneck_2021}
Uri Alon and Eran Yahav.
\newblock On the {Bottleneck} of {Graph} {Neural} {Networks} and its {Practical} {Implications}.
\newblock In \emph{International {Conference} on {Learning} {Representations}}, 2021.

\bibitem[Ambrosetti et~al.(2014)Ambrosetti, Reilly, DiStasio, and Tkatchenko]{ambrosetti_long-range_2014}
Alberto Ambrosetti, Anthony~M. Reilly, Robert~A. DiStasio, Jr., and Alexandre Tkatchenko.
\newblock Long-range correlation energy calculated from coupled atomic response functions.
\newblock \emph{The Journal of Chemical Physics}, 140\penalty0 (18):\penalty0 18A508, February 2014.

\bibitem[Arnaiz-Rodríguez et~al.(2022)Arnaiz-Rodríguez, Begga, Escolano, and Oliver]{arnaiz-rodriguez_diffwire_2022}
Adrián Arnaiz-Rodríguez, Ahmed Begga, Francisco Escolano, and Nuria~M. Oliver.
\newblock {DiffWire}: {Inductive} {Graph} {Rewiring} via the {Lovász} {Bound}.
\newblock In \emph{Learning on {Graphs} {Conference}}, November 2022.

\bibitem[Balcilar et~al.(2020)Balcilar, Renton, Héroux, Gaüzère, Adam, and Honeine]{balcilar_analyzing_2020}
Muhammet Balcilar, Guillaume Renton, Pierre Héroux, Benoit Gaüzère, Sébastien Adam, and Paul Honeine.
\newblock Analyzing the {Expressive} {Power} of {Graph} {Neural} {Networks} in a {Spectral} {Perspective}.
\newblock In \emph{International {Conference} on {Learning} {Representations}}, October 2020.

\bibitem[Bastos et~al.(2023)Bastos, Nadgeri, Singh, Suzumura, and Singh]{bastos_learnable_2023}
Anson Bastos, Abhishek Nadgeri, Kuldeep Singh, Toyotaro Suzumura, and Manish Singh.
\newblock Learnable {Spectral} {Wavelets} on {Dynamic} {Graphs} to {Capture} {Global} {Interactions}.
\newblock In \emph{Proceedings of the {Thirty}-{Seventh} {AAAI} {Conference} on {Artificial} {Intelligence}}, volume~37 of \emph{{AAAI}'23/{IAAI}'23/{EAAI}'23}, pp.\  6779--6787. AAAI Press, February 2023.

\bibitem[Bo et~al.(2023)Bo, Shi, Wang, and Liao]{bo_specformer_2023}
Deyu Bo, Chuan Shi, Lele Wang, and Renjie Liao.
\newblock Specformer: {Spectral} {Graph} {Neural} {Networks} {Meet} {Transformers}.
\newblock In \emph{The {Eleventh} {International} {Conference} on {Learning} {Representations}}, February 2023.

\bibitem[Bracewell(2000)]{bracewell_fourier_2000}
Ronald~N. Bracewell.
\newblock \emph{The {Fourier} transform and its applications}.
\newblock {McGraw}-{Hill} series in electrical and computer engineering {Circuits} and systems. McGraw-Hill, Boston, 3. ed edition, 2000.

\bibitem[Bronstein et~al.(2017)Bronstein, Bruna, LeCun, Szlam, and Vandergheynst]{bronstein_geometric_2017}
Michael~M. Bronstein, Joan Bruna, Yann LeCun, Arthur Szlam, and Pierre Vandergheynst.
\newblock Geometric {Deep} {Learning}: {Going} beyond {Euclidean} data.
\newblock \emph{IEEE Signal Processing Magazine}, 34\penalty0 (4):\penalty0 18--42, July 2017.
\newblock Conference Name: IEEE Signal Processing Magazine.

\bibitem[Cho et~al.(2024)Cho, Sim, Wu, and Kim]{cho_neurodegenerative_2024}
Hyuna Cho, Jaeyoon Sim, Guorong Wu, and Won~Hwa Kim.
\newblock Neurodegenerative {Brain} {Network} {Classification} via {Adaptive} {Diffusion} with {Temporal} {Regularization}.
\newblock In \emph{Forty-first {International} {Conference} on {Machine} {Learning}}, June 2024.

\bibitem[Chung(1997)]{chung_spectral_1997}
Fan R.~K. Chung.
\newblock \emph{Spectral {Graph} {Theory}}.
\newblock American Mathematical Soc., 1997.

\bibitem[Defferrard et~al.(2016)Defferrard, Bresson, and Vandergheynst]{defferrard_convolutional_2016}
Michaël Defferrard, Xavier Bresson, and Pierre Vandergheynst.
\newblock Convolutional {Neural} {Networks} on {Graphs} with {Fast} {Localized} {Spectral} {Filtering}.
\newblock In \emph{Advances in {Neural} {Information} {Processing} {Systems}}, volume~29, pp.\  3844--3852. Curran Associates, Inc., 2016.

\bibitem[Dokholyan(2016)]{dokholyan_controlling_2016}
Nikolay~V. Dokholyan.
\newblock Controlling {Allosteric} {Networks} in {Proteins}.
\newblock \emph{Chemical Reviews}, 116\penalty0 (11):\penalty0 6463--6487, June 2016.
\newblock Publisher: American Chemical Society.

\bibitem[Dwivedi et~al.(2021)Dwivedi, Luu, Laurent, Bengio, and Bresson]{dwivedi_graph_2021}
Vijay~Prakash Dwivedi, Anh~Tuan Luu, Thomas Laurent, Yoshua Bengio, and Xavier Bresson.
\newblock Graph {Neural} {Networks} with {Learnable} {Structural} and {Positional} {Representations}.
\newblock In \emph{International {Conference} on {Learning} {Representations}}, October 2021.

\bibitem[Dwivedi et~al.(2022)Dwivedi, Rampášek, Galkin, Parviz, Wolf, Luu, and Beaini]{dwivedi_long_2022}
Vijay~Prakash Dwivedi, Ladislav Rampášek, Michael Galkin, Ali Parviz, Guy Wolf, Anh~Tuan Luu, and Dominique Beaini.
\newblock Long {Range} {Graph} {Benchmark}.
\newblock \emph{Advances in Neural Information Processing Systems}, 35:\penalty0 22326--22340, December 2022.

\bibitem[Dwivedi et~al.(2023)Dwivedi, Joshi, Luu, Laurent, Bengio, and Bresson]{dwivedi_benchmarking_2023}
Vijay~Prakash Dwivedi, Chaitanya~K. Joshi, Anh~Tuan Luu, Thomas Laurent, Yoshua Bengio, and Xavier Bresson.
\newblock Benchmarking {Graph} {Neural} {Networks}.
\newblock \emph{Journal of Machine Learning Research}, 24\penalty0 (43):\penalty0 1--48, 2023.

\bibitem[Gasteiger et~al.(2019)Gasteiger, Weiß~enberger, and Günnemann]{gasteiger_diffusion_2019}
Johannes Gasteiger, Stefan Weiß~enberger, and Stephan Günnemann.
\newblock Diffusion {Improves} {Graph} {Learning}.
\newblock In \emph{Advances in {Neural} {Information} {Processing} {Systems}}, volume~32, pp.\  13354--13366. Curran Associates, Inc., 2019.

\bibitem[Geisler et~al.(2023)Geisler, Li, Mankowitz, Cemgil, Günnemann, and Paduraru]{geisler_transformers_2023}
Simon Geisler, Yujia Li, Daniel~J. Mankowitz, Ali~Taylan Cemgil, Stephan Günnemann, and Cosmin Paduraru.
\newblock Transformers {Meet} {Directed} {Graphs}.
\newblock In \emph{Proceedings of the 40th {International} {Conference} on {Machine} {Learning}}, pp.\  11144--11172. PMLR, July 2023.
\newblock ISSN: 2640-3498.

\bibitem[Geisler et~al.(2024)Geisler, Kosmala, Herbst, and Günnemann]{geisler2024spatiospectral}
Simon Geisler, Arthur Kosmala, Daniel Herbst, and Stephan Günnemann.
\newblock Spatio-{Spectral} {Graph} {Neural} {Networks}, June 2024.
\newblock arXiv:2405.19121 [cs].

\bibitem[Gilmer et~al.(2017)Gilmer, Schoenholz, Riley, Vinyals, and Dahl]{gilmer_neural_2017}
Justin Gilmer, Samuel~S. Schoenholz, Patrick~F. Riley, Oriol Vinyals, and George~E. Dahl.
\newblock Neural {Message} {Passing} for {Quantum} {Chemistry}.
\newblock In \emph{Proceedings of the 34th {International} {Conference} on {Machine} {Learning}}, pp.\  1263--1272. PMLR, July 2017.

\bibitem[Glorot \& Bengio(2010)Glorot and Bengio]{glorot_understanding_2010}
Xavier Glorot and Yoshua Bengio.
\newblock Understanding the difficulty of training deep feedforward neural networks.
\newblock In \emph{Proceedings of the {Thirteenth} {International} {Conference} on {Artificial} {Intelligence} and {Statistics}}, pp.\  249--256. JMLR Workshop and Conference Proceedings, March 2010.
\newblock ISSN: 1938-7228.

\bibitem[Gori et~al.(2005)Gori, Monfardini, and Scarselli]{gori_new_2005}
Marco Gori, Gabriele Monfardini, and Franco Scarselli.
\newblock A {New} {Model} for {Learning} in {Graph} {Domains}.
\newblock In \emph{Proceedings. 2005 {IEEE} {International} {Joint} {Conference} on {Neural} {Networks}}, volume~2, pp.\  729--734, July 2005.

\bibitem[Hammond et~al.(2011)Hammond, Vandergheynst, and Gribonval]{hammond_wavelets_2011}
David~K. Hammond, Pierre Vandergheynst, and Rémi Gribonval.
\newblock Wavelets on graphs via spectral graph theory.
\newblock \emph{Applied and Computational Harmonic Analysis}, 30\penalty0 (2):\penalty0 129--150, March 2011.

\bibitem[He et~al.(2021)He, Wei, Huang, and Xu]{he_bernnet_2021}
Mingguo He, Zhewei Wei, Zengfeng Huang, and Hongteng Xu.
\newblock {BernNet}: {Learning} {Arbitrary} {Graph} {Spectral} {Filters} via {Bernstein} {Approximation}.
\newblock In \emph{Advances in {Neural} {Information} {Processing} {Systems}}, November 2021.

\bibitem[Kipf \& Welling(2016)Kipf and Welling]{kipf_semi-supervised_2016}
Thomas~N. Kipf and Max Welling.
\newblock Semi-{Supervised} {Classification} with {Graph} {Convolutional} {Networks}.
\newblock In \emph{International {Conference} on {Learning} {Representations}}, November 2016.

\bibitem[Knörzer et~al.(2022)Knörzer, van Diepen, Hsiao, Giedke, Mukhopadhyay, Reichl, Wegscheider, Cirac, and Vandersypen]{knorzer_long-range_2022}
J.~Knörzer, C.~J. van Diepen, T.-K. Hsiao, G.~Giedke, U.~Mukhopadhyay, C.~Reichl, W.~Wegscheider, J.~I. Cirac, and L.~M.~K. Vandersypen.
\newblock Long-range electron-electron interactions in quantum dot systems and applications in quantum chemistry.
\newblock \emph{Physical Review Research}, 4\penalty0 (3):\penalty0 033043, July 2022.
\newblock Publisher: American Physical Society.

\bibitem[Kreuzer et~al.(2021)Kreuzer, Beaini, Hamilton, Létourneau, and Tossou]{kreuzer_rethinking_2021}
Devin Kreuzer, Dominique Beaini, William~L. Hamilton, Vincent Létourneau, and Prudencio Tossou.
\newblock Rethinking {Graph} {Transformers} with {Spectral} {Attention}.
\newblock In \emph{Advances in {Neural} {Information} {Processing} {Systems}}, volume~34. Curran Associates, Inc., November 2021.

\bibitem[Liao et~al.(2018)Liao, Zhao, Urtasun, and Zemel]{liao_lanczosnet_2018}
Renjie Liao, Zhizhen Zhao, Raquel Urtasun, and Richard Zemel.
\newblock {LanczosNet}: {Multi}-{Scale} {Deep} {Graph} {Convolutional} {Networks}.
\newblock In \emph{International {Conference} on {Learning} {Representations}}, September 2018.

\bibitem[Liu et~al.(2024{\natexlab{a}})Liu, He, Laurent, Di~Giovanni, Bronstein, and Bresson]{liu_advancing_2024}
Nian Liu, Xiaoxin He, Thomas Laurent, Francesco Di~Giovanni, Michael~M. Bronstein, and Xavier Bresson.
\newblock Advancing {Graph} {Convolutional} {Networks} via {General} {Spectral} {Wavelets}, May 2024{\natexlab{a}}.
\newblock arXiv:2405.13806 [cs].

\bibitem[Liu et~al.(2025)Liu, He, Laurent, Giovanni, Bronstein, and Bresson]{liu_general_2025}
Nian Liu, Xiaoxin He, Thomas Laurent, Francesco~Di Giovanni, Michael~M. Bronstein, and Xavier Bresson.
\newblock A {General} {Graph} {Spectral} {Wavelet} {Convolution} via {Chebyshev} {Order} {Decomposition}, May 2025.
\newblock arXiv:2405.13806 [cs].

\bibitem[Liu et~al.(2024{\natexlab{b}})Liu, Yin, Liu, and Wang]{liu_aswt-sgnn_2024}
Ruyue Liu, Rong Yin, Yong Liu, and Weiping Wang.
\newblock {ASWT}-{SGNN}: {Adaptive} {Spectral} {Wavelet} {Transform}-{Based} {Self}-{Supervised} {Graph} {Neural} {Network}.
\newblock \emph{Proceedings of the AAAI Conference on Artificial Intelligence}, 38\penalty0 (12):\penalty0 13990--13998, March 2024{\natexlab{b}}.
\newblock Number: 12.

\bibitem[Loshchilov \& Hutter(2017)Loshchilov and Hutter]{loshchilov_sgdr_2017}
Ilya Loshchilov and Frank Hutter.
\newblock {SGDR}: {Stochastic} {Gradient} {Descent} with {Warm} {Restarts}.
\newblock In \emph{International {Conference} on {Learning} {Representations}}, February 2017.

\bibitem[Loshchilov \& Hutter(2018)Loshchilov and Hutter]{loshchilov_decoupled_2018}
Ilya Loshchilov and Frank Hutter.
\newblock Decoupled {Weight} {Decay} {Regularization}.
\newblock In \emph{International {Conference} on {Learning} {Representations}}, September 2018.

\bibitem[Mallat(1989)]{mallat_multiresolution_1989}
Stephane~G. Mallat.
\newblock Multiresolution {Approximations} and {Wavelet} {Orthonormal} {Bases} of {L} 2 ({R}).
\newblock \emph{Transactions of the American Mathematical Society}, 315\penalty0 (1):\penalty0 69, September 1989.

\bibitem[Mallat(2009)]{mallat_wavelet_2009}
Stéphane Mallat.
\newblock \emph{A {Wavelet} {Tour} of {Signal} {Processing}}.
\newblock Academic Press, Boston, 3rd edition, 2009.

\bibitem[Markov(1889)]{markov_question_1889}
Andrei~Andreyevich Markov.
\newblock On a question by {D}.{I}. {Mendeleev}.
\newblock \emph{Zapiski Imperatorskoi Akademii Nauk}, 62:\penalty0 1--24, October 1889.

\bibitem[McAuley et~al.(2015)McAuley, Targett, Shi, and van~den Hengel]{mcauley_image-based_2015}
Julian McAuley, Christopher Targett, Qinfeng Shi, and Anton van~den Hengel.
\newblock Image-{Based} {Recommendations} on {Styles} and {Substitutes}.
\newblock In \emph{Proceedings of the 38th {International} {ACM} {SIGIR} {Conference} on {Research} and {Development} in {Information} {Retrieval}}, {SIGIR} '15, pp.\  43--52, New York, NY, USA, August 2015. Association for Computing Machinery.

\bibitem[Min et~al.(2022)Min, Chen, Bian, Xu, Zhao, Huang, Zhao, Huang, Ananiadou, and Rong]{min_transformer_2022}
Erxue Min, Runfa Chen, Yatao Bian, Tingyang Xu, Kangfei Zhao, Wenbing Huang, Peilin Zhao, Junzhou Huang, Sophia Ananiadou, and Yu~Rong.
\newblock Transformer for {Graphs}: {An} {Overview} from {Architecture} {Perspective}.
\newblock \emph{arXiv preprint arXiv:2202.08455}, February 2022.
\newblock arXiv:2202.08455 [cs].

\bibitem[Rampášek et~al.(2023)Rampášek, Galkin, Dwivedi, Luu, Wolf, and Beaini]{rampasek_recipe_2023}
Ladislav Rampášek, Mikhail Galkin, Vijay~Prakash Dwivedi, Anh~Tuan Luu, Guy Wolf, and Dominique Beaini.
\newblock Recipe for a {General}, {Powerful}, {Scalable} {Graph} {Transformer}.
\newblock \emph{arXiv preprint arXiv:2205.12454}, January 2023.
\newblock arXiv:2205.12454 [cs].

\bibitem[Sandryhaila \& Moura(2013)Sandryhaila and Moura]{sandryhaila_discrete_2013}
Aliaksei Sandryhaila and José M.~F. Moura.
\newblock Discrete {Signal} {Processing} on {Graphs}.
\newblock \emph{IEEE Transactions on Signal Processing}, 61\penalty0 (7):\penalty0 1644--1656, April 2013.
\newblock Conference Name: IEEE Transactions on Signal Processing.

\bibitem[Scarselli et~al.(2009)Scarselli, Gori, {Ah Chung Tsoi}, Hagenbuchner, and Monfardini]{scarselli_graph_2009}
Franco Scarselli, Marco Gori, {Ah Chung Tsoi}, Markus Hagenbuchner, and Gabriele Monfardini.
\newblock The {Graph} {Neural} {Network} {Model}.
\newblock \emph{IEEE Transactions on Neural Networks}, 20\penalty0 (1):\penalty0 61--80, January 2009.
\newblock Conference Name: IEEE Transactions on Neural Networks.

\bibitem[Schütt et~al.(2017)Schütt, Kindermans, Sauceda~Felix, Chmiela, Tkatchenko, and Müller]{schutt_schnet_2017}
Kristof Schütt, Pieter-Jan Kindermans, Huziel~Enoc Sauceda~Felix, Stefan Chmiela, Alexandre Tkatchenko, and Klaus-Robert Müller.
\newblock {SchNet}: {A} continuous-filter convolutional neural network for modeling quantum interactions.
\newblock In \emph{Advances in {Neural} {Information} {Processing} {Systems}}, volume~30. Curran Associates, Inc., 2017.

\bibitem[Shchur et~al.(2018)Shchur, Mumme, Bojchevski, and Günnemann]{shchur_pitfalls_2018}
Oleksandr Shchur, Maximilian Mumme, Aleksandar Bojchevski, and Stephan Günnemann.
\newblock Pitfalls of {Graph} {Neural} {Network} {Evaluation}.
\newblock In \emph{Relational {Representation} {Learning} {Workshop}, {Advances} in {Neural} {Information} {Processing} {Systems}}, volume~32, November 2018.

\bibitem[Shuman et~al.(2011)Shuman, Vandergheynst, and Frossard]{shuman_chebyshev_2011}
David~I Shuman, Pierre Vandergheynst, and Pascal Frossard.
\newblock Chebyshev polynomial approximation for distributed signal processing.
\newblock In \emph{2011 {International} {Conference} on {Distributed} {Computing} in {Sensor} {Systems} and {Workshops} ({DCOSS})}, pp.\  1--8, June 2011.

\bibitem[Stachenfeld et~al.(2020)Stachenfeld, Godwin, and Battaglia]{stachenfeld_graph_2020}
Kimberly Stachenfeld, Jonathan Godwin, and Peter Battaglia.
\newblock Graph {Networks} with {Spectral} {Message} {Passing}, December 2020.
\newblock arXiv:2101.00079 [stat].

\bibitem[Tönshoff et~al.(2023)Tönshoff, Ritzert, Rosenbluth, and Grohe]{tonshoff_where_2023}
Jan Tönshoff, Martin Ritzert, Eran Rosenbluth, and Martin Grohe.
\newblock Where {Did} the {Gap} {Go}? {Reassessing} the {Long}-{Range} {Graph} {Benchmark}.
\newblock In \emph{The {Second} {Learning} on {Graphs} {Conference}}, November 2023.

\bibitem[Wang \& Zhang(2022)Wang and Zhang]{wang_how_2022}
Xiyuan Wang and Muhan Zhang.
\newblock How {Powerful} are {Spectral} {Graph} {Neural} {Networks}.
\newblock In \emph{Proceedings of the 39th {International} {Conference} on {Machine} {Learning}}, pp.\  23341--23362. PMLR, June 2022.

\bibitem[Xu et~al.(2019)Xu, Shen, Cao, Qiu, and Cheng]{xu_graph_2019}
Bingbing Xu, Huawei Shen, Qi~Cao, Yunqi Qiu, and Xueqi Cheng.
\newblock Graph {Wavelet} {Neural} {Network}.
\newblock In \emph{International {Conference} on {Learning} {Representations}}, 2019.

\bibitem[Zhu et~al.(2022)Zhu, Wang, Han, and Xu]{zhu_neural_2022}
Jingxuan Zhu, Juexin Wang, Weiwei Han, and Dong Xu.
\newblock Neural relational inference to learn long-range allosteric interactions in proteins from molecular dynamics simulations.
\newblock \emph{Nature Communications}, 13\penalty0 (1):\penalty0 1661, March 2022.
\newblock Publisher: Nature Publishing Group.

\end{thebibliography}
